%% file: paper.tex
\documentclass[runningheads]{llncs}
\usepackage{graphicx}

\usepackage{tikz}
\usepackage{comment}
\usepackage{amsmath,amssymb} 
\usepackage{color}

\usepackage[accsupp]{axessibility}  

\usepackage{multirow}
\usepackage{subfiles}
\usepackage{pgfplots}
\usepackage{makecell}
\usepackage[colorlinks,linkcolor=red,anchorcolor=blue,citecolor=green]{hyperref}

\newcommand{\ECCVSubNumber}{7230}  

\newcommand{\BIB}{./ref.bib}

\newcommand{\MARK}[1]{\textcolor{black}{\textbf{#1}}}

\newcommand{\SOCIALGAN}{S-GAN}
\newcommand{\SOPHIE}{SoPhie}
\newcommand{\BIGAT}{Social-BiGAT}
\newcommand{\SIMAUG}{SimAug}
\newcommand{\PECNET}{PECNet}

\newcommand{\SRPAMI}{E-SR-LSTM}
\newcommand{\GARDEN}{Multiverse}

\newcommand{\TRANSFORMER}{STAR}
\newcommand{\TP}{TF}
\newcommand{\TPNMS}{TPNMS}
\newcommand{\TRAJECTRONPP}{Trajectron++}
\newcommand{\INTROVERT}{Introvert}
\newcommand{\LBEBM}{LB-EBM}
\newcommand{\YNET}{Y-net}
\newcommand{\SPECTGNN}{SpecTGNN}

\newcommand{\AGENTFORMER}{Agentformer}
\newcommand{\MANTRA}{MANTRA}

\newcommand{\MODEL}{V$^{2}$-Net}

\newcommand{\FIG}[1]{\figurename\ref{#1}}
\newcommand{\TABLE}[1]{Table \ref{#1}}
\newcommand{\EQUA}[1]{Equation \ref{#1}}

\newcommand{\xmark}{$\times$}
\newcommand{\xm}{\xmark}
\newcommand{\cm}{\checkmark}
\newcommand{\SOTA}{state-of-the-art}
\newcommand{\ADE}{ADE}
\newcommand{\FDE}{FDE}
\newcommand{\ADEFULL}{Average Displacement Error}
\newcommand{\FDEFULL}{Final Displacement Error}
\newcommand{\ETAL}{{\emph{et al.}}}
\newcommand{\EG}{{\emph{e.g.}}}
\newcommand{\IE}{{\emph{i.e.}}}
\newcommand{\SUBTITLE}[1]{\subsubsection{#1.}}

\newcommand \footnoteONLYtext[1]{
	\let \mybackup \thefootnote
	\let \thefootnote \relax
	\footnotetext{#1}
	\let \thefootnote \mybackup
	\let \mybackup \imareallyundefinedcommand
}

\begin{document}
\pagestyle{headings}
\mainmatter

\title{View Vertically: A Hierarchical Network for Trajectory Prediction via Fourier Spectrums} 


\titlerunning{\MODEL~for Trajectory Prediction}
%
\author{
    Conghao Wong* \inst{1} \and
    Beihao Xia* \inst{1} \and
    Ziming Hong \inst{1} \and
    Qinmu Peng \inst{1} \and
    Wei Yuan \inst{1} \and
    Qiong Cao \inst{2} \and
    Yibo Yang \inst{2} \and
    Xinge You \inst{1}
}%

\footnoteONLYtext{* Equal contribution.}
\footnoteONLYtext{~~Codes are available at \url{https://github.com/cocoon2wong/Vertical}.}

\authorrunning{C. Wong, B. Xia, et al.}
%
\institute{
    Huazhong University of Science and Technology \and
    JD Explore Academy \\
    \email{\{conghao\_wong, xbh\_hust\}@hust.edu.cn}
}

\maketitle

\subfile{./contents/abstract.tex}

\subfile{./contents/introduction.tex}

\subfile{./contents/relatedWork.tex}

\subfile{./contents/method.tex}

\subfile{./contents/experiments.tex}

\clearpage
%
%
\bibliographystyle{splncs04}
\bibliography{\BIB}

\subfile{./contents/appendix.tex}

\end{document}

%% file: contents/abstract.tex
\begin{abstract}
    Understanding and forecasting future trajectories of agents are critical for behavior analysis, robot navigation, autonomous cars, and other related applications.
    Previous methods mostly treat trajectory prediction as time sequence generation.
    Different from them, this work studies agents' trajectories in a ``vertical'' view, \IE, modeling and forecasting trajectories from the spectral domain.
    Different frequency bands in the trajectory spectrums could hierarchically reflect agents' motion preferences at different scales.
    The low-frequency and high-frequency portions could represent their coarse motion trends and fine motion variations, respectively.
    Accordingly, we propose a hierarchical network V$^2$-Net, which contains two sub-networks, to hierarchically model and predict agents' trajectories with trajectory spectrums.
    The coarse-level keypoints estimation sub-network first predicts the ``minimal'' spectrums of agents' trajectories on several ``key'' frequency portions.
    Then the fine-level spectrum interpolation sub-network interpolates the spectrums to reconstruct the final predictions.
    Experimental results display the competitiveness and superiority of V$^2$-Net on both ETH-UCY benchmark and the Stanford Drone Dataset.

    \keywords{Hierarchical Trajectory Prediction, Fourier Spectrums}

\end{abstract}

%% file: contents/introduction.tex
\section{Introduction}

\subfile{./_fig_intro.tex}

Trajectory prediction aims at inferring agents' possible future trajectories considering potential influencing factors.
It has been an essential but challenging task, which can be widely applied to behavior analysis \cite{alahi2017learning,chai2019multipath,Phan-Minh_2020_CVPR}, robot navigation \cite{unfreezing}, autonomous driving \cite{cui2019multimodal,deo2018convolutional,kim2017probabilistic,desire}, tracking \cite{youWillNeverWalkAlone,saleh2020artist}, detection \cite{2016Anomaly,fernando2018soft+,mehran2009abnormal}, and many other computer vision tasks \cite{morris2011trajectory,xie2017learning}.
Researchers have widely studied interactive factors, including the agent-agent interaction \cite{socialLSTM,socialGAN} and the agent-scene interaction \cite{bigat,sophie}.
Another line of researchers have explored creative ways to model trajectories.
Neural networks like Long-Short Term Memory Networks (LSTMs) \cite{hug2018particle,manh2018scene,ssLSTM}, Graph Convolution Networks (GCNs) \cite{Li_2021_ICCV,Sun_2020_CVPR,xue2020scene}, and Transformers \cite{giuliari2020transformer,yu2020spatio} are employed to encode agents' trajectories.
Some researchers have also studied agents' multi-modal characteristics \cite{chai2019multipath,trajectron} and the distributions of their goals \cite{choi2019drogon,girase2021loki,rhinehart2019precog}, thus forecasting multiple future choices.

However, most previous methods treat trajectory prediction as time sequence generation and predict trajectories recurrently, which could be challenging to reflect agents' motion preferences at different scales hierarchically.
In other words, researchers usually tend to explore agents' behaviors and their changing trends dynamically but lack the overall analyses at different temporal scales.
In fact, pedestrians always plan their activities at different levels simultaneously.
For example, they may first determine their coarse motion trends and then make fine decisions to interactive variations.
Although some methods \cite{yu2020spatio,giuliari2020transformer,Yuan_2021_ICCV} employ neural networks with attention mechanisms (like Transformers) as the backbone to model agents' status, they may still be difficult to directly represent agents' detailed motion differences at different temporal scales.

The Fourier transform (FT) and its variations have significantly succeeded in the signal processing community.
Recently, researchers have also introduced FTs to some computer vision tasks, such as image de-noising \cite{8296825}, edge extraction \cite{KAUR2021107852}, and image super-resolution \cite{CHENG201921}.
FTs decompose sequential inputs into a series of sinusoids with different amplitudes and phases on different frequencies.
Furthermore, these sinusoids could reflect the differentiated frequency response characteristics at different frequency scales, which could be difficult to obtain directly in the original signal.
They provide a ``vertical'' view for processing and analyzing sequences, thus presenting elusive features in the original signals.

Some researchers have applied FTs in tasks similar to trajectory prediction.
For example, Mao \ETAL~\cite{mao2019learning,mao2020history} employ the discrete cosine transform (DCT) to help predict human motions.
Cao \ETAL~\cite{cao2021spectral} use the graph Fourier transform to attempt to model interaction series in trajectory prediction.
Unfortunately, FTs have not been employed to \emph{model trajectories} in the trajectory prediction task directly.
Inspired by the successful use of FTs, we try to employ the Discrete Fourier Transform (DFT) to obtain trajectory spectrums to capture agents' detailed motion preferences at different frequency scales.
\FIG{fig_intro}(a) demonstrates an agent's two-dimensional trajectory in the time series and spectrum view, respectively.
Although the two projection trajectories have similar shapes, they show considerable differences in the spectrum view.
It means that the trajectory spectrums (including amplitudes and phases) obtained through FTs could reflect subtle differences that are difficult to discern within the time series.

Furthermore, some works have divided trajectory prediction into a two-stage pipeline, which we call the \emph{hierarchical} prediction strategy.
\cite{tran2021goal,mangalam2020not} divide this two-stage process into destination prediction and destination-controlled prediction.
\cite{mangalam2020s} introduces several ``waypoints'' to help predict agents' potential intentions rather than the only destination point.
FTs decompose time series into the combination of different frequency portions.
Inspired by these hierarchical approaches, a natural thought is to hierarchically predict agents' trajectories on different \emph{frequency scales}, including:
(a) \emph{Global Plannings}, \IE, agents' coarse motion trends.
The low-frequency portions (slow-changing portions) in the trajectory spectrums could reflect their plannings globally.
(b) \emph{Interactive Preferences}, \IE, agents' detailed interactions.
The high-frequency portions (fast-changing portions) in the spectrums will directly show the rapidly changing movement rules, thus further describing their personalized interactive preferences.

\FIG{fig_intro}(b) demonstrates the trajectory reconstructed by different number of frequency portions.
Agent's overall plannings could be reconstructed through a few low-frequency portions of the trajectory spectrum.
By continually adding new high-frequency portions, the reconstructed trajectory would be able to reflect finer motion details and interactive preferences.
Accordingly, we introduce a ``coarse-fine'' strategy to hierarchically model global plannings and interactive preferences by trajectory spectrums at different levels correspondingly.

In this paper, we propose the \MODEL~to hierarchically forecast trajectories in the spectrum view.
It contains two sub-networks.
The coarse-level keypoints estimation sub-network first predicts the ``minimal'' spectrums of agents' trajectories on several ``key'' frequency portions, and then the fine-level spectrum interpolation sub-network interpolates the spectrums to reconstruct the final predictions.
Meanwhile, agents' social and scene interactions will also be concerned to make the network available to give predictions that conform to social rules and physical constraints.
Our contributions are list as follows:
\begin{itemize}
    \item We introduce the Fourier Transform to model and predict trajectories with spectrums to better capture agents' behaviors from a different perspective.
    \item A Transformer-based \MODEL~containing coarse-level keypoints estimation and fine-level spectrum interpolation sub-networks is proposed to model and predict trajectories ``coarse-to-fine'' hierarchically.
    \item Experiments demonstrate that \MODEL~achieves competitive performance on both ETH-UCY benchmark and the Stanford Drone Dataset.
\end{itemize}

%% file: contents/_fig_intro.tex
\begin{figure}
    \centering
    \includegraphics[width=1.0\linewidth]{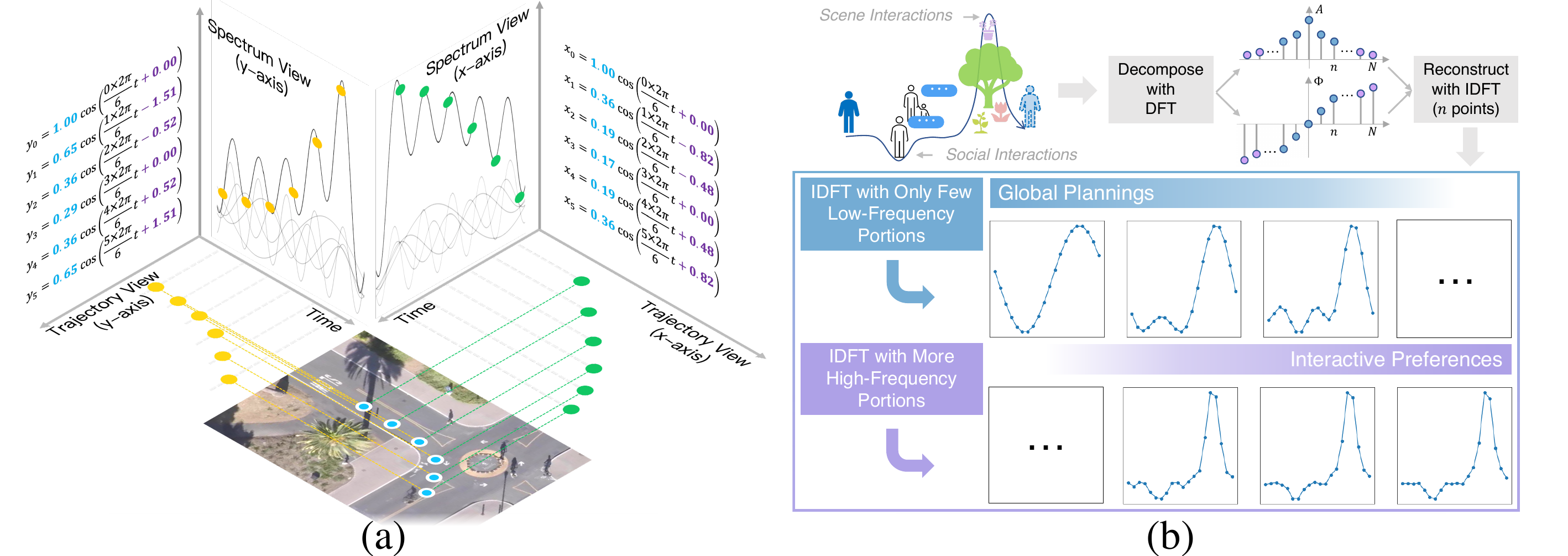}
    \caption{
        \MODEL~Motivation.
        (a) We show the trajectory in two views, the time sequences view and the Fourier spectrums view.
        The x-axis trajectory (green dots) and the y-axis trajectory (yellow dots) have similar ``shapes'', while they are quite different in the spectrum views (shown with amplitudes and phases), which motivates us to view trajectories ``vertically'' with spectrums rather than time sequences.
        (b) shows a trajectory with interactions.
        We utilize the first $n/N$ low-frequency portions to reconstruct the trajectory, and the results show that different frequency portions have different contributions to their global plannings and interactive preferences.
    }
    \label{fig_intro}
\end{figure}

%% file: contents/relatedWork.tex
\section{Related Work}

\SUBTITLE{Trajectory Prediction}

Recently, researchers have studied how interactive factors affect agents' trajectory plans, like social interaction \cite{socialLSTM,stgat,stgcnn,srLSTM} and scene interaction \cite{manh2018scene,ssLSTM}.
The modeling of agents' trajectories also matters how the trajectory prediction networks perform.
Alahi \ETAL~\cite{socialLSTM} treat this task as a sequence generation problem, and they employ LSTMs to model and predict pedestrians' positions in the next time step recurrently.
A series of works have also introduced Graph Neural Networks (GNNs), \EG~Graph Attention Networks (GATs) \cite{stgat,bigat,Liang_2020_CVPR} and GCNs \cite{cao2020spectral,stgcnn}, to handle interactive factors when forecasting.
Moreover, the attention mechanism has been employed \cite{socialAttention,cidnn,srLSTM} to focus on the most valuable interactive targets to give reasonable predictions.
With the success of Transformers \cite{attentionIsAllYouNeed} in sequence processing such as natural language processing, researchers \cite{giuliari2020transformer,yu2020spatio} have employed Transformers to obtain better feature representations.
Some researchers address agents' uncertainty and randomness by introducing generative models.
Generative Adversarial Networks (GANs) are employed in \cite{socialGAN,sophie,bigat} to generate multiple stochastic trajectories to suit agents' various future choices.
Some works like \cite{trajectron,salzmann2020trajectron} use the Conditional Variation AutoEncoder (CVAE) to achieve the similar goal.

\SUBTITLE{Applications of Fourier Transforms}

Many approaches achieve better performance in different computer vision tasks by introducing FTs.
Cheng \ETAL~\cite{CHENG201921} present a Fast Fourier Transform-based algorithm, which brings high computational efficiency and reliability for the multichannel interpolation in image super-resolution.
Kaur \ETAL~\cite{KAUR2021107852} propose a Fractional Fourier Transform based Riesz fractional derivative approach for edge detection and apply it to enhance images.
Komatsu \ETAL~\cite{8296825} construct the 3-D mean-separation-type short-time DFT and apply it to denoise moving images.
It is worth noting that FTs have been widely applied in handling time-series forecasting problems.
Mao \ETAL~\cite{mao2020history,mao2019learning} employ DCT to help predict human skeleton-graphs in motion forecasting.
Cao \ETAL~\cite{cao2020spectral,cao2021spectral} propose spectral temporal graph to model interaction series (not trajectories) in trajectory forecasting.
Forecasting trajectories could also be regarded as one of the time-series forecasting tasks \cite{socialLSTM}.
The successful use of FTs motivates us to model \emph{agents' trajectories} in the Fourier domain, therefore trying to obtain better representations.
Unfortunately, there seems to be no method that directly uses FTs to describe agents' trajectories in the field of trajectory prediction.
This paper attempts to model and forecast \emph{agents' trajectories} in the spectrum view for the first time.

\SUBTITLE{Hierarchical Trajectory Prediction}

More and more researchers \cite{choi2019drogon,girase2021loki,rhinehart2019precog} have treated trajectory prediction as a ``two-stage'' problem and predict trajectories \emph{hierarchically}.
Some researchers divide this process into ``destination prediction'' and ``destination-controlled prediction''.
Mangalam \ETAL~\cite{mangalam2020not} infer trajectory endpoints to assist in long-range multi-modal trajectory prediction.
Tran \ETAL~\cite{tran2021goal} attempt to obtain multi-modal goal proposals from the additional goal channel to generate multiple predictions.
Wong \ETAL~\cite{wong2021msn} use a set of generators to give destination proposals with different styles, and then interpolate to forecast multiple future predictions.
Others like \cite{mangalam2020s} also introduce several ``waypoints'' to help better predict agents' potential future intentions rather than the only destination points.
These hierarchical approaches concern more on factor ``destination'' (or ``waypoints'') and ``predictions''.
Considering the Fourier transforms could decompose time series into the combination of different frequency portions, a natural thought is to predict agents' future trajectories hierarchically on different \emph{frequency scales}.
Therefore, these two-stage factors may become the ``low-frequency-portions'' and ``high-frequency-portions'' for trajectory spectrums.
It motivates us to forecast trajectories from coarse global plannings to fine interactive preferences with trajectory spectrums hierarchically.

%% file: contents/method.tex
\section{Method}

\subfile{_fig_overview.tex}

\subfile{_tab_method.tex}

As shown in \FIG{fig_overview}, \MODEL~has two main sub-networks, the coarse-level keypoints estimation sub-network and the fine-level spectrum interpolation sub-network.

\SUBTITLE{Formulations}

Let $p_t = (x_t, y_t) \in \mathbb{R}^2$ denote the two-dimensional coordinates of one agent at step $t$.
Given a video clip $\{I\}$ that contains $M$ agents' observed trajectories $\{X_i\}_{i=1}^M$ ($X_i = (p_1, p_2, ..., p_{t_h})^T$ represents $i$-th agent's observed trajectory) during the observation period, trajectory prediction aims to forecast their possible future coordinates $\{\hat{Y}_i\}_{i=1}^{M}$ ($\hat{Y}_i = (\hat{p}_{t_h + 1}, \hat{p}_{t_h + 2}, ..., \hat{p}_{t_h + t_f})^T$ denotes one of the prediction) during the corresponding future $t_f$ steps considering their observations and the interactive context.

\SUBTITLE{Keypoints Estimation Sub-network}

The coarse-level keypoints estimation sub-network aims to forecast agents' keypoint spectrums with \emph{a lower spatio-temporal resolution}.
An encoder-decoder structure is designed to generate multiple random predictions to adapt to agents' uncertainty and randomness.

We first use Discrete Fourier Transform (DFT) on agents' observed trajectories to obtain the $t_h$-point trajectory spectrums.
We apply 1D-DFT on \emph{each dimension} in the observed trajectory $X = \left\{({x_t}, {y_t})\right\}_{t=1}^{t_h}$ to obtain their spectrums, including the amplitudes $A = \{a_x, a_y\}$ and the phases $\Phi = \{\phi_x, \phi_y\}$:
\begin{equation}
    \label{eq_1}
    \begin{aligned}
        \mathcal{X} & = \mbox{DFT}[({x_1}, {x_2}, ..., {x_{t_h}})] \in \mathbb{C}^{t_h}, ~~
        \mathcal{Y} = \mbox{DFT}[({y_1}, {y_2}, ..., {y_{t_h}})] \in \mathbb{C}^{t_h},          \\
        A           & = \{a_x, a_y\} = \{\Vert \mathcal{X} \Vert, \Vert \mathcal{Y} \Vert\}, ~~
        \Phi = \{\phi_x, \phi_y\} = \{\arg \mathcal{X}, \arg \mathcal{Y}\}.
    \end{aligned}
\end{equation}

We employ an embedding MLP (the MLP$_t$ in \TABLE{tab_method}) to embed agents' observed trajectory spectrums $(a_x, a_y, \phi_x, \phi_y) \in \mathbb{R}^{t_h \times 4}$ into the high-dimensional $f_t$.
Similar to previous works like \SOCIALGAN~\cite{socialGAN}, we model agents' multimodality by sampling the random noise vector $z \sim \mathcal{N}(0, I)$ and then concatenating the corresponding random representations $f_i$ to the $f_t$.
The encoder for these noise vectors (we call the MLP$_i$) has the same structure like the MLP$_t$.
We combine the above representations to obtain the embedded vector $f_e$.
Formally,
\begin{equation}
    \label{eq_2}
    \begin{aligned}
        f_t & = \mbox{MLP}_t((a_x, a_y, \phi_x, \phi_y)) \in \mathbb{R}^{t_h \times 64}, \\
        f_i & = \mbox{MLP}_i(z) \in \mathbb{R}^{t_h \times 64},                          \\
        f_e & = [f_t, f_i] \in \mathbb{R}^{t_h \times 128}.
    \end{aligned}
\end{equation}
Here, $[a, b]$ represents the concatenation for vectors $\{a, b\}$ on the last dimension.

Then, we use a Transformer\cite{attentionIsAllYouNeed} \footnote{Please see Appendix for the Transformer details.} named T$_{k}$ to encode agents' behavior representations.
The embedded vector $f_e$ is passed to the Transformer encoder, and the spectrums $(a_x, a_y, \phi_x, \phi_y)$ are input to the Transformer decoder.
The Transformer here is used as the feature extractor, and it does not contain the final output layer.
We employ another MLP encoder (MLP$_e$) to aggregate features at different frequency nodes \cite{wong2021msn}, thus inferring the behavior feature $f$:
\begin{equation}
    \label{eq_3}
    \begin{aligned}
        f & = \mbox{MLP}_e \left( \mbox{T}_{k}(f_e, (a_x, a_y, \phi_x, \phi_y)) \right) \in \mathbb{R}^{t_h \times 128}.
    \end{aligned}
\end{equation}

The sub-network finally utilizes a decoder MLP (MLP$_{d}$) to predict agents' $N_{key}$-point ($N_{key} < t_f$) \textbf{keypoint spectrums} $[A^{key}, \Phi^{key}]$.
Formally,
\begin{equation}
    \label{eq_keypointsprediction}
    [A^{key}, \Phi^{key}] = (a_{x}^{key}, a_{y}^{key}, \phi_{x}^{key}, \phi_{y}^{key}) = \mbox{MLP}_{d}(f) \in \mathbb{R}^{N_{key} \times 4}.
\end{equation}
We call $N_{key}$ \textbf{the number of spectrum keypoints}.

When training the sub-network, agents' key spatial coordinates $y^{key}$ (which are gathered from their entire ground-truth future trajectories) will be used as the supervision.
Meanwhile, the Inverse DFT (IDFT) will be applied to obtain the predicted spatial keypoints $\hat{y}^{key}$.
The network variables will be tuned by minimizing the average Euclidean distance between $y^{key}$ and the predicted $\hat{y}^{key}$, therefore learning to predict the corresponding keypoint spectrums.
We define the \textbf{Average Keypoints Loss} ($\mathcal{L}_{\mbox{AKL}}$) as:
\begin{equation}
    \mathcal{L}_{\mbox{AKL}} = \Vert \hat{y}^{key} - y^{key} \Vert_2 = \frac{1}{N_{key}} \sum_{n=1}^{N_{key}} \Vert \hat{p}_{t_n^{key}} - p_{t_n^{key}} \Vert_2,
\end{equation}
where
\begin{equation}
    \begin{aligned}
        \hat{y}^{key} & = \left(\mbox{IDFT}[a_{x}^{key} \exp(j\phi_{x}^{key})],~\mbox{IDFT}[a_{y}^{key} \exp(j\phi_{y}^{key})]\right) \in \mathbb{R}^{N_{key} \times 2}, \\
        y^{key}       & = \left(p_{t^{key}_{1}}, p_{t^{key}_{2}}, ..., p_{t^{key}_{N_{key}}}\right)^T \in \mathbb{R}^{N_{key} \times 2}.
    \end{aligned}
\end{equation}

\SUBTITLE{Spectrum Interpolation Sub-network}

The fine-level spectrum interpolation sub-network reconstructs the complete trajectory spectrums from the keypoint spectrums with \emph{a higher spatio-temporal resolution}.
It aims to learn the spectrum biases between the complete spectrums and the keypoint spectrums.

The sub-network takes the $N_{key}$-point keypoint spectrums $[A^{key}, \Phi^{key}]$ as the input.
Similar to \EQUA{eq_2}, we have the representation $f^{key}_t$:
\begin{equation}
    \label{eq_keydft}
    f^{key}_t = \mbox{MLP}_t((a_{x}^{key}, a_{y}^{key}, \phi_{x}^{key}, \phi_{y}^{key})) \in \mathbb{R}^{N_{key} \times 64}.
\end{equation}
Note that MLP$_t$ in \EQUA{eq_2} and \EQUA{eq_keydft} do not share weights.

Besides, agents' social and scene interactions will be concerned in this sub-network.
In detail, we use a context embedding MLP (MLP$_c$) to encode the transferred images $C$ \cite{xia2022cscnet} (which encodes both social interactions and scene constraints together in an energy map form by the scene visual image $I$ and trajectories $\mathcal{T}$ via a CNN) into the context feature $f_c = \mbox{MLP}_c(C)$.

We employ a similar Transformer (called the Interpolation Transformer, T$_i$) to learn the spectrum biases.
We pass the concatenated feature $f_e^{key} = [f_t^{key}, f_c]$ to the transformer encoder, and keypoint spectrums $[A^{key}, \Phi^{key}] = (a_{x}^{key}, a_{y}^{key}, \phi_{x}^{key}, \phi_{y}^{key})$ to the transformer decoder.
The transformer here is used to forecast the \textbf{complete spectrums} $[\hat{A}, \hat{\Phi}] = (\hat{a}_x, \hat{a}_y, \hat{\phi}_x, \hat{\phi}_y)$.
Then, we use the IDFT to obtain the reconstructed trajectory $\hat{y}_o$.
Formally,
\begin{equation}
    \label{eq_6}
    \begin{aligned}
         & (\hat{a}_x, \hat{a}_y, \hat{\phi}_x, \hat{\phi}_y) = \mbox{T}_i(f_e^{key}, (a_{x}^{key}, a_{y}^{key}, \phi_{x}^{key}, \phi_{y}^{key})) \in \mathbb{R}^{(t_h + t_f) \times 4}, \\
         & \hat{y}_o = \left(\mbox{IDFT}[\hat{a}_x \exp(j\hat{\phi}_x)],~\mbox{IDFT}[\hat{a}_y \exp(j\hat{\phi}_y)]\right) \in \mathbb{R}^{(t_h + t_f) \times 2}.
    \end{aligned}
\end{equation}
Finally, we have one of the \MODEL~predictions:
\begin{equation}
    \hat{y} = \hat{y}_o [t_h:, :] \in \mathbb{R}^{t_f \times 2},
\end{equation}
where the $[t_h:, :]$ indicates the slicing operation on tensors.

The spectrum interpolation sub-network learns to interpolate the ``key'' trajectory spectrums into the complete trajectory spectrums, thus reflecting agents' fine interactive details by predicting the remaining spectrum portions.
Its variables will be tuned through the \textbf{Average Point-wise Loss} ($\mathcal{L}_{\mbox{APL}}$):
\begin{equation}
    \mathcal{L}_{\mbox{APL}} = \Vert \hat{y} - y \Vert_2 = \frac{1}{t_f} \sum_{t=t_h + 1}^{t_h + t_f} \Vert \hat{p}_{t} - p_{t} \Vert_2.
\end{equation}

\SUBTITLE{Loss Functions}

We use the joint loss function to tain \MODEL:
\begin{equation}
    \begin{aligned}
        \mathcal{L}= \mathcal{L}_{\mbox{AKL}} + \mathcal{L}_{\mbox{APL}}.
    \end{aligned}
\end{equation}

%% file: contents/_fig_overview.tex
\begin{figure*}[tbp]
    \centering
    \includegraphics[width=1.0\linewidth]{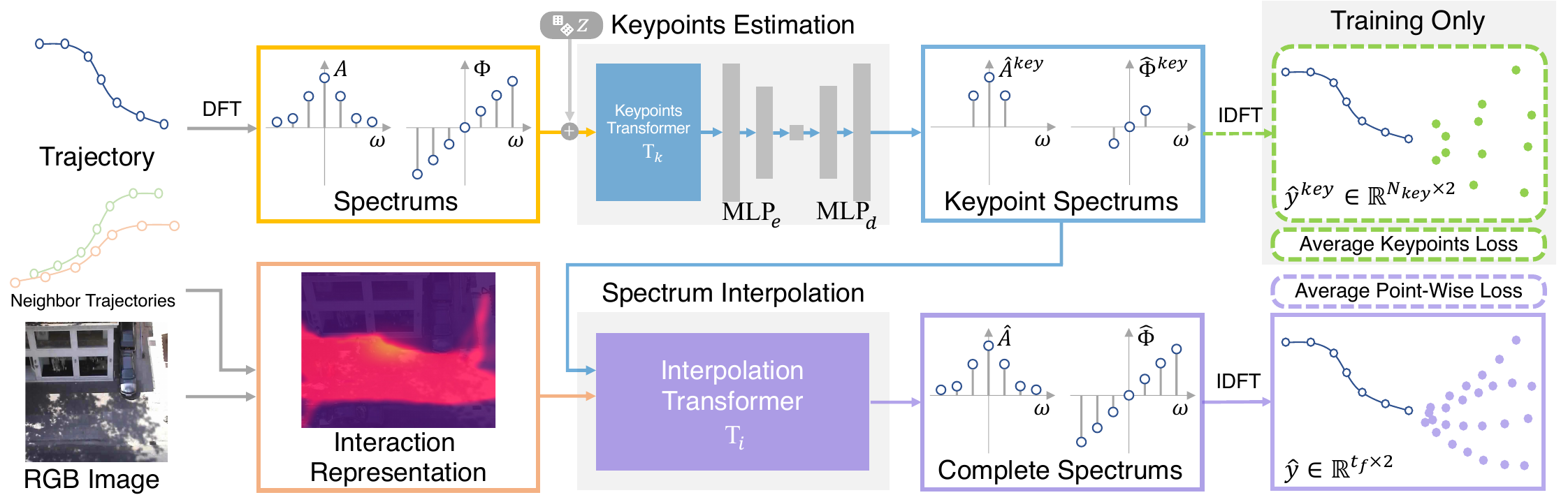}
    \caption{
        \MODEL~Overview.
        The coarse-level keypoints estimation sub-network aims at forecasting the $N_{key}$-point keypoint spectrums $[A^{key}, \Phi^{key}]$ from the observed trajectory spectrums $[A, \Phi]$.
        The other fine-level spectrum interpolation sub-network aims at predicting the complete spectrums $[\hat{A}, \hat{\Phi}]$ on each future moment based on the keypoint spectrums and the interaction representations.
    }
    \label{fig_overview}
\end{figure*}

%% file: contents/_tab_method.tex
\begin{table}[tbp]
    \centering
    \caption{
        Architecture details of the proposed \MODEL.
    }
    \label{tab_method}
    \begin{tabular}{c|c}
        \hline
        Layers & Network Architecture \\
        
        \hline
        MLP$_t$ & \makecell[l]{
            $(a_x, a_y, \phi_x, \phi_y) \to \mbox{fc}(64, \mbox{ReLU})$ 
            $\to \mbox{fc}(64, \tanh) \to f_t$ 
        }\\

        \hline
        MLP$_c$ & \makecell[l]{
            $C \to$ MaxPool$(5 \times 5) \to$ Flatten 
            $\to$ $\mbox{fc}(64t_h, \tanh)$ $\to$ Reshape($t_h , 64$) $\to f_c$
        }\\

        \hline
        T$_k$ & \makecell[l]{
            $f_e$ $\to$ TransformerEncoder ($128$) $\to f_k'$;\\
            $f_k', (a_x, a_y, \phi_x, \phi_y)$ $\to$ 
            TransformerDecoder ($128$) $\to f_k''$\\
        }\\

        \hline
        MLP$_e$ & \makecell[l]{
            $f_k'' \to $ fc($128, \tanh$) $\to$ fc($128$) $\to f$
        }\\

        \hline
        MLP$_{d}$ & \makecell[l]{
            $f \to$ $\mbox{fc}(128, \mbox{ReLU}) \to \mbox{fc}(128, \mbox{ReLU})$ 
            $\to \mbox{fc}(4N_{key}) \to $ \\
            $ \mbox{Reshape}~(N_{key} , 4)$ $\to (a_x^{key},a_y^{key},\phi_x^{key},\phi_y^{key})$
        }\\

        \hline
        T$_i$ & \makecell[l]{
            $[f_t^{key}, f_c] \to$ TransformerEncoder ($128$) $\to f_i'$;\\
            $f_i', (a_{x}^{key}, a_{y}^{key}, \phi_{x}^{key}, \phi_{y}^{key}) \to$
            TransformerDecoder ($4$) $\to (\hat{a}_x, \hat{a}_y, \hat{\phi}_x, \hat{\phi}_y)$
        }\\
        
        \hline
    \end{tabular}
\end{table}

%% file: contents/experiments.tex
\section{Experiments}

\SUBTITLE{Datasets}

(a) \textbf{ETH-UCY Benchmark}:
Many previous methods like \cite{socialLSTM,socialGAN,sophie} take several sub-datasets from ETH\cite{youWillNeverWalkAlone} and UCY\cite{2007Crowds} to train and evaluate their models with the ``leave-one-out'' strategy \cite{socialLSTM}, which is called the ETH-UCY benchmark.
It contains 1536 pedestrians with thousands of non-linear trajectories.
The annotations are pedestrians' coordinates in meters.
(b) \textbf{Stanford Drone Dataset}:
The Stanford Drone Dataset \cite{learningSocialEtiquette} (SDD) has 60 bird-view videos captured by drones.
More than 11,000 different agents are annotated with bounding boxes in pixels.
It contains over 185,000 social interactions and 40,000 scene interactions, which is more complex and challenging.
\footnote{
    Dataset splits used to train and validate on SDD are the same as \cite{liang2020simaug}.
}

\SUBTITLE{Metrics}

We employ two metrics to evaluate the prediction accuracy, including the \ADEFULL~(\ADE) and the \FDEFULL~(\FDE) \cite{socialLSTM,youWillNeverWalkAlone}.
\ADE~is the average point-wise Euclidean distance between groundtruth and predictions of all steps, and \FDE~is the Euclidean distance between the last point's prediction and groundtruth.
Following \cite{socialGAN}, the reported metrics are the minimum value in 20 generations (``best-of-20'').
For one prediction, we have:
\begin{equation}
    \begin{aligned}
        \mbox{\ADE} = \frac{1}{t_f} \sum_{t = t_h + 1}^{t_h + t_f} \Vert p_t - \hat{p}_t \Vert_2,~~
        \mbox{\FDE} = \Vert p_{t_h + t_f} - \hat{p}_{t_h + t_f} \Vert_2.
    \end{aligned}
\end{equation}

\SUBTITLE{Baselines}

We choose several methods as our baselines, including
\SOCIALGAN~\cite{socialGAN},
\SOPHIE~\cite{sophie},
\BIGAT~\cite{bigat},
\SRPAMI~\cite{zhang2020social},
\MANTRA~\cite{Marchetti_2020_CVPR},
\GARDEN~\cite{Liang_2020_CVPR},
\SIMAUG~\cite{liang2020simaug},
\PECNET~\cite{mangalam2020not},
\TRANSFORMER~\cite{yu2020spatio}
\TPNMS~\cite{liang2020temporal},
\TP~\cite{giuliari2020transformer},
\TRAJECTRONPP~\cite{salzmann2020trajectron},
\INTROVERT~\cite{shafiee2021introvert},
\LBEBM~\cite{pang2021trajectory},
\AGENTFORMER~\cite{Yuan_2021_ICCV},
\YNET~\cite{mangalam2020s}, and
\SPECTGNN~\cite{cao2021spectral}.

\SUBTITLE{Implementation Details}

We predict agents' trajectories in future $t_f = 12$ frames according to their $t_h = 8$ frames' observations.
The frame rate is set to 2.5 frames per second when sampling trajectories.
We train the entire \MODEL~with the Adam optimizer (learning rate lr $=0.0003$) on one NVIDIA Tesla P4 graphic processor.
\MODEL~is trained with the batch size bs $=2500$ for 800 epochs on ETH-UCY and 150 epochs on SDD.
Detailed layer connections, output units, and activations are listed in \TABLE{tab_method}.
We employ $L = 4$ layers of encoder-decoder structure with $H = 8$ attention heads in each Transformer-based sub-network.
The output dimension of fully connected layers used in multi-head attention layers is set to 128.
The default number of spectrum keypoints is set to $N_{key} = 3$.
When training the network, we set $\{t^{key}_1, t^{key}_2, t^{key}_3\} = \{t_{h} + 4, t_{h} + 8, t_{h} + 12\}$.

\subsection{Comparision to State-of-the-Art Methods}

\SUBTITLE{ETH-UCY}

As shown in \TABLE{tab_ade_all}, \MODEL~achieves better performance on most ETH-UCY sub-datasets.
It improves the metrics on eth by 23.3\% and 28.8\% compared with \LBEBM.
Meanwhile, \MODEL~has better performance than \INTROVERT~on hotel by about 8.8\% \FDE.
Compared with \INTROVERT, its average \ADE~and \FDE~have been improved by 14.2\% and 17.6\%, respectively.
Moreover, \MODEL~has achieved comparable average performance to the \SOTA~\AGENTFORMER~and \YNET.
It shows \MODEL's strong competitiveness on ETH-UCY.
Especially, \MODEL~dramatically outperforms \AGENTFORMER~and \YNET~about 14.8\% ADE on eth, which demonstrates its unique advantages.

\SUBTITLE{SDD}

As shown in \TABLE{tab_ade_sdd}, \MODEL~has better performance than the previous \SOTA~methods on SDD.
Compared to \SPECTGNN, \MODEL~has significantly gained by 13.3\% and 8.2\% in \ADE~and \FDE, respectively.
\MODEL~also outperforms the \SOTA~\YNET~by 9.3\% and 3.9\%.
In general, \MODEL~has a better prediction accuracy than other baselines on SDD.
It is worth noting that SDD is more complex than ETH-UCY, which demonstrates \MODEL's robustness and adaptability to larger and more challenging scenarios.

\subfile{_tab_ade_all.tex}

\subfile{_tab_ab.tex}

\subsection{Quantitative Analysis}

We design several model variations and run ablation studies to verify each design in the proposed \MODEL.
Quantitative results are shown in \TABLE{tab_ab} and \TABLE{tab_ab_dft}.

\SUBTITLE{DFT on Trajectories}

\MODEL~implements on the trajectory spectrums to focus on different frequency portions in different prediction stages.
Variations a1 and a2 are trained with the same number of spectrum keypoints $N_{key} = 1$, while a1 does not use DFT/IDFT when predicting.
It means that variation a2 obtains behavior features and interaction features from the spectrums, while a1 obtains them from the time series.
Experiments show that a2 outperforms a1 by about 10\% average \ADE~and \FDE~on ETH-UCY.
In addition, we remove all other components from the model and design two minimal model variations min1 and min2 to demonstrate more directly the usefulness of introducing DFT for trajectory prediction.
These two variations DO NOT take into account agents' interactions and multimodal properties, but only retain the Transformer backbone to achieve the ``simple'' trajectory prediction goal.
We train both the two variations under the same conditions.
Results in \TABLE{tab_ab_dft} favorably demonstrate the performance gains (12.9\% on ADE and 11.1\% on FDE) by using trajectory spectrums, independent of the rest of the designs in the proposed network.
It proves the effectiveness of DFT in this task, and further demonstrates the considerable performance gain by implementing on the trajectory spectrums.

\SUBTITLE{Number of Spectrum Keypoints $N_{key}$}

The number of spectrum keypoints to be predicted matters how the keypoints estimation sub-network determines agents' overall motion trends.
A smaller $N_{key}$ might cause a looser planning division, making it challenging to reflect the differences between agents' similar future choices.
On the contrary, a larger value may lead to a strict trends division, which could be difficult for the subsequent network to reflect agents' multiple uncertain future choices and motion preferences.
Ablation experiments on variations a2, b1, b2, and b3 show the quantitative comparisons with different $N_{key}$ configurations.
The temporal keypoints are set to the equivalence point among the predicted period ($t_f = 12$).
For example, when $N_{key} = 1$, we set $\{t_1^{key}\} = \{t_h + 12\}$, and when $N_{key} = 4$, we set $\{t_1^{key}, t_2^{key}, t_3^{key}, t_4^{key}\} = \{t_h + 3, t_h + 6, t_h + 9, t_h + 12\}$.
It shows that variation b1 ($N_{key} = 3$) outperforms variation a2 ($N_{key} = 1$) with additional improvements by about 5\% \ADE~and 6\% \FDE.
However, The performance of variation b3 ($N_{key} = 6$) reduces by about 5\% and 1\% compared to variation a1.
Please refer to visualized trajectories of different $N_{key}$ variations in section ``Qualitative Analysis'' to see how they affect the prediction performance qualitatively.

\SUBTITLE{Hierarchical Prediction}

Similar to some \emph{goal-driven methods}, \MODEL~splits trajectory prediction into a two-stage ``keypoints-interpolation'' process.
However, the most significant difference from these methods is that our model implements the entire process with trajectory spectrums rather than spatial coordinates.
Correspondingly, the second stage ``goal-conditioned prediction'' has turned into the ``spectrum interpolation''.
Although the first stage sub-network has already given a low-resolution prediction composed of a few keypoint spectrums, it is not easy to reflect agents' subtle interactions and activity differences.
The spectrum interpolation sub-network aims to interpolate these ``key'' spectrums into the complete trajectory spectrums, thus reflecting agents' fine-level interactive preferences.
Variation c only implements the S1 sub-network but utilizes linear interpolation to finish forecasting.
Variations b1 and c could show the improvements brought by the interpolation sub-network.
Results point out that b1 improves about 7\% in ADE compared with variation c.
It effectively reflects the superiority of the ``coarse-to-fine'' hierarchical prediction with spectrums.

\subfile{_fig_keypoints.tex}

\subsection{Qualitative Analysis}

\SUBTITLE{Coarse-Fine Prediction}

\MODEL~predicts agents' multiple random keypoint spectrums at the first stage, and then uses a spectrum interpolation method to forecast the complete spectrums by considering interactions and fine activities.
\FIG{fig_keypoints} shows two visualized results at keypoints estimation and spectrum interpolation stages correspondingly.
At the first stage, \MODEL~aims to predict keypoint spectrums with a low spatio-temporal resolution.
We apply IDFT on these keypoint spectrums to show the corresponding spatial keypoints in \FIG{fig_keypoints}.
As shown in \FIG{fig_keypoints}(a) and (b), these spatial keypoints could only represent the general trend of agents' future activities rather than the specific and precise coordinates.
Then, the interpolation sub-network considers both physical constraints and social interactions to ``fine-tune'' these keypoints.
For example, the marked keypoints in \FIG{fig_keypoints}(b) indicate that the pedestrian could go toward the road.
However, the corresponding final prediction shows that he might slow down and stay on the side of the road considering the scene's physical constraints.

\subfile{_fig_number_keypoints.tex}

\SUBTITLE{Number of Spectrum Keypoints and Prediction Styles}

The keypoint sub-network estimate agents' keypoint spectrums to determine their coarse future routes.
Meanwhile, it contains a generative module to give agents multiple predictions by randomly sampling $z \sim N(0, I)$, therefore adapting to agents' uncertain future decisions.
The number of spectrum keypoints $N_{key}$ matters directly \MODEL's stochastic performance, which can be seen in \FIG{fig_number_keypoints}.
It shows that \FIG{fig_number_keypoints}(a) exhibits a looser restraint when generating trajectories.
Obviously, it is caused by the insufficient trends constraints from the little $N_{key}$.
Although (a) gives a large number of future predictions, most of them may be invalid or do not meet the scene constraints.
Visualized results indicate that \MODEL~will give more rigorous predictions when $N_{key}$ increases.
It will forecast more acceptable predictions when increasing the $N_{key}$ to 3 (\FIG{fig_number_keypoints}(b)).
However, its predictions will be limited to a small area when $N_{key}$ is set to 6 (\FIG{fig_number_keypoints}(c)).
At this time, \MODEL~could hardly reflect agents' multiple stochastic plannings under the constraints of such excessive keypoints.
In short, with a low number of keypoints a wider spectrum of possibilities is covered (higher multimodality) and vice versa by increasing the number the trajectories become more plausible (collisions with static elements are avoided).
It also explains the drop of quantitative performance when $N_{key} = 6$ in the above ablation studies.
Furthermore, we can set different $N_{key}$ in different prediction scenarios to achieve controllable predictions.

\subfile{_fig_vis.tex}

\SUBTITLE{Visualization}

As shown in \FIG{fig_vis}, \MODEL~could predict multiple trajectories considering both scene constraints and social interactions.
For example, \MODEL~gives a variety of different future options for bikers crossing the intersection (illustrated in \FIG{fig_vis}(a)(e)(k)).
In the prediction case (e) and (k), it considers the physical constraints of the environment and forecast to bypasses the grass.
Additionally, it also shows strong adaptability in some unique prediction scenarios.
For instance, it gives three kinds of predictions to the biker in case (h) to pass through the traffic circle: turning right, going ahead, and turning left.
When turning left, the given predictions are not to turn left directly (just like turning right) but to go left after going around the traffic circle, which is consistent with the traffic rules around the circle.
Case (i) presents similar results for the man passing the crossroads.
It shows \MODEL~could accurately describe the scene's physical constraints and agents' motion rules in various scenes.

Surprisingly, \MODEL~could give ``smoother'' predictions due to the usage of DFT, like turning left in case (h).
The spatio-temporal continuity between adjacent points in the predicted time series will be ensured with the superposition of a series of sinusoids.
Therefore, the predictions could reflect the real physical characteristics of the target agent while keeping their interactive preferences.

\subfile{_fig_failure.tex}

\SUBTITLE{Limitations}

As shown in \FIG{fig_failure}, \MODEL~has failed predictions for stationary agents and few interactive cases.
Although not all predictions are regarded as failed predictions, some of them show not-so-reasonable movement trends for these standing still agents.
Additionally, it may provide predictions not that suitable for the interactive context.
We will fix this problem in the future.

\section{Conclusion}
In this work, we focus on giving predictions by modeling their trajectories with spectrums from the global plannings and the interactive preferences two hierarchical steps.
A Transformer-based \MODEL, which consists of the coarse-level keypoints estimation sub-network and the fine-level spectrum interpolation sub-network, is proposed to predict agents' trajectories hierarchically.
Experimental results show that \MODEL~ achieves the competitive performance on both ETH-UCY benchmark and the Stanford Drone Dataset.
Although the proposed method shows higher prediction accuracy and provides better visualized results, there are still some failed predictions.
We will address this problem and further explore feasible solutions to model and predict agents' possible trajectories.

\subsubsection{Acknowledgements.}

This work was partially supported by the National Natural Science Foundation of China (Grant No. 62172177), and in part by the Fundamental Research Funds for the Central Universities (Grant No. 2021yjsCXCY040).

%% file: contents/_tab_ade_all.tex
\begin{table*}[tbp]
    \centering
    \caption{
        Comparisons to baselines with the \emph{best-of-20} on ETH-UCY.
        Metrics are shown in the format of ``\ADE/\FDE'' in meters.
        Lower means better.
    }
    \label{tab_ade_all}
    \begin{tabular}{c|ccccc|c}
        \hline
        Models & eth & hotel & univ & zara1 & zara2 & Average \\
    
        \hline
        \BIGAT~\cite{bigat} & 0.69/1.29 & 0.49/1.01 & 0.55/1.32 & 0.30/0.62 & 0.36/0.75 & 0.48/1.00 \\
        \SRPAMI~\cite{zhang2020social} & 0.44/0.79 & 0.19/0.31 & 0.50/1.05 & 0.32/0.64 & 0.27/0.54 & 0.34/0.67 \\
        \PECNET~\cite{mangalam2020not} & 0.54/0.87 & 0.18/0.24 & 0.35/0.60 & 0.22/0.39 & 0.17/0.30 & 0.29/0.48 \\
        \TRANSFORMER~\cite{yu2020spatio} & 0.56/1.11 & 0.26/0.50 & 0.52/1.13 & 0.40/0.89 & 0.52/1.13 & 0.41/0.87 \\
        \TPNMS~\cite{liang2020temporal} & 0.52/0.89 & 0.22/0.39 & 0.55/0.13 & 0.35/0.70 & 0.27/0.56 & 0.38/0.73 \\
        \TP~\cite{giuliari2020transformer} & 0.61/1.12 & 0.18/0.30 & 0.35/0.65 & 0.22/0.38 & 0.17/0.32 & 0.31/0.55 \\
        \TRAJECTRONPP~\cite{salzmann2020trajectron} & 0.43/0.86 & 0.12/0.19 & \MARK{0.22}/0.43 & 0.17/0.32 & \MARK{0.12}/0.25 & \MARK{0.20}/0.39 \\
        
        \INTROVERT~\cite{shafiee2021introvert} & 0.42/0.70 & 0.11/0.17 & \MARK{0.20}/\MARK{0.32} & \MARK{0.16}/0.27 & 0.16/0.25 & 0.21/0.34 \\
        \LBEBM~\cite{pang2021trajectory} & 0.30/0.52 & 0.13/0.20 & 0.27/0.52 & 0.20/0.37 & 0.15/0.29 & 0.21/0.38 \\
        \AGENTFORMER~\cite{Yuan_2021_ICCV} & 0.26/0.39 & 0.11/\MARK{0.14} & 0.26/0.46 & \MARK{0.15/0.23} & \MARK{0.14/0.23} & \MARK{0.18/0.29} \\

        \YNET~\cite{mangalam2020s} & 0.28/\MARK{0.33} & \MARK{0.10}/\MARK{0.14} & 0.24/0.41 & 0.17/0.27 & \MARK{0.13/0.22} & \MARK{0.18}/\MARK{0.27} \\
        \hline
        \MODEL~(Ours) 
        & \MARK{0.23}/0.37 
        & \MARK{0.11/0.16} 
        & \MARK{0.21}/0.35 
        & 0.19/0.30 
        & \MARK{0.14/0.24} 
        & \MARK{0.18/0.28}\\
        
        \hline
    \end{tabular}
\end{table*}

\begin{table*}[tbp]
    \centering
    \caption{
        Comparisons to baselines with the \emph{best-of-20} on SDD.
        Lower means better.
    }
    \label{tab_ade_sdd}
    \begin{tabular}{c|c|c|c|c|c}
        \hline
        Models & 
        \SOCIALGAN~\cite{socialGAN} &
        \SOPHIE~\cite{sophie} &
        \GARDEN~\cite{Liang_2020_CVPR} &
        \SIMAUG~\cite{liang2020simaug} &
        \PECNET \\ 
        \hline
        ADE/FDE & 
        27.25/41.44 & 
        16.27/29.38 & 
        14.78/27.09 & 
        12.03/23.98 & 
        9.96/15.88 \\
        \hline
        \hline
        Models & 
        \MANTRA~\cite{Marchetti_2020_CVPR} &
        \LBEBM &
        \SPECTGNN~\cite{cao2021spectral} &
        \YNET &
        \MODEL~(Ours) \\
        \hline
        ADE/FDE & 
        8.96/17.76 &
        8.87/15.61 & 
        8.21/12.41 & 
        7.85/11.85 &
        \MARK{7.12/11.39}

\\


        \hline
    \end{tabular}
    
\end{table*}

%% file: contents/_tab_ab.tex
\begin{table*}[btp]
    \centering
    \caption{
        Ablation Studies.
        S1 and S2 represent the keypoints estimation stage and the spectrum interpolation stage, correspondingly.
        ``D'' indicates whether different variations model trajectories with spectrums.
        ``N'' indicates the number of spectrum keypoints.
        Results in ``$\uparrow$'' are the percentage metric improvements compared to a1.
    }
    \label{tab_ab}
    \begin{tabular}{c|cccc|p{1.45cm}<{\centering}|p{1.45cm}<{\centering}|p{1.45cm}<{\centering}|p{1.45cm}<{\centering}|p{1.45cm}<{\centering}|c}
        \hline
        No. & S1 & S2 & D & N  &
        eth & hotel & univ & zara1 & zara2 & $\uparrow$ Gain (\%) 
        \\

        \hline
        a1 & \cm & \cm & \xm & 1 &
        0.29/0.47 & 0.12/0.18 & 0.23/0.38 & 0.25/0.37 & 0.17/0.29 
        & - (base) 
        \\
        
        a2 & \cm & \cm & \cm & 1 & 
        0.23/0.39 & 0.12/0.17 & 0.23/0.37 & 0.22/0.32 & 0.16/0.26 
        & 7.9\%/11.3\% 
        \\

        \hline
        b1 & \cm & \cm & \cm & 3 &
        0.23/0.37 & 0.11/0.16 & 0.21/0.35 & 0.19/0.30 & 0.14/0.24 
        & 16.2\%/16.5\% 
        \\

        b2 & \cm & \cm & \cm & 4 &
        0.24/0.38 & 0.12/0.17 & 0.22/0.36 & 0.21/0.32 & 0.16/0.26 
        & 9.5\%/12.4\% 
        \\

        b3 & \cm & \cm & \cm & 6 &
        0.29/0.49 & 0.13/0.17 & 0.23/0.38 & 0.26/0.37 & 0.21/0.31 
        & -6.0\%/-1.2\% 
        \\

        \hline
        c & \cm & \xm & \cm & 3 &
        0.24/0.37 & 0.12/0.16 & 0.23/0.35 & 0.21/0.30 & 0.15/0.25 
        & 9.6\%/15.9\% 
        \\
        
        \hline
    \end{tabular}
    
\end{table*}

\begin{table*}[btp]
    \centering
    \caption{
        Validation of DFT.
        We design and run two minimal model variations to show the performance gain brought by DFT.
        Symbols are the same as \TABLE{tab_ab}.
    }
    \label{tab_ab_dft}
    \begin{tabular}{c|cccc|p{1.45cm}<{\centering}|p{1.45cm}<{\centering}|p{1.45cm}<{\centering}|p{1.45cm}<{\centering}|p{1.45cm}<{\centering}|c}
        \hline
        No. & S1 & S2 & D & N  &
        eth & hotel & univ & zara1 & zara2 & $\uparrow$ Gain (\%) 
        \\

        \hline
        min1 & \cm & \xm & \xm & - & 0.83/1.66 & 0.25/0.44 & 0.77/1.39 & 0.48/0.97 & 0.38/0.74 & - (base) 
        \\

        min2 & \cm & \xm & \cm & - & 0.79/1.51 & 0.22/0.38 & 0.55/1.10 & 0.46/0.92 & 0.34/0.71 & 12.9\%/11.1\%
        \\
        
        \hline
    \end{tabular}
    
\end{table*}

%% file: contents/_fig_keypoints.tex
\begin{figure}[tbp]
    \centering
    \includegraphics[width=1.0\linewidth]{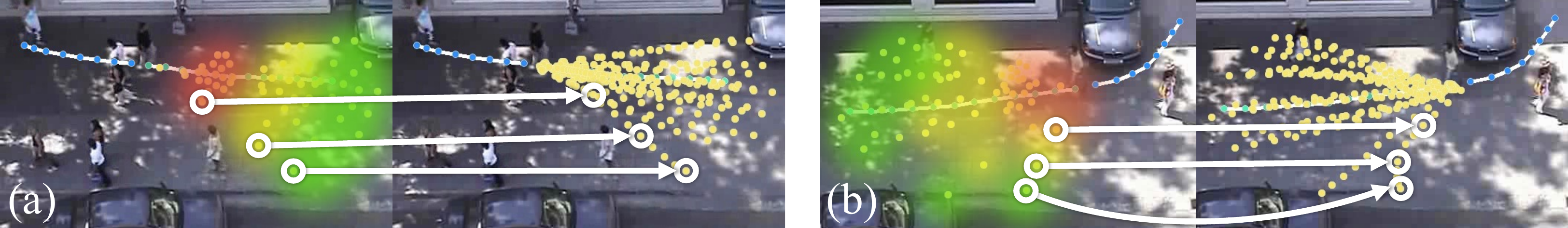}
    \caption{
        Keypoints and Interpolation Illustration.
        We show the predicted spatial keypoints ($N_{key} = 3$) after IDFT and the corresponding predicted trajectories \MODEL~finally outputs.
        Their corresponding time steps are distinguished with different color masks.
        Dots connected by white arrows belong to the same prediction steps.
    }
    \label{fig_keypoints}
\end{figure}

%% file: contents/_fig_number_keypoints.tex
\begin{figure}[tbp]
    \centering
    \includegraphics[width=1.0\linewidth]{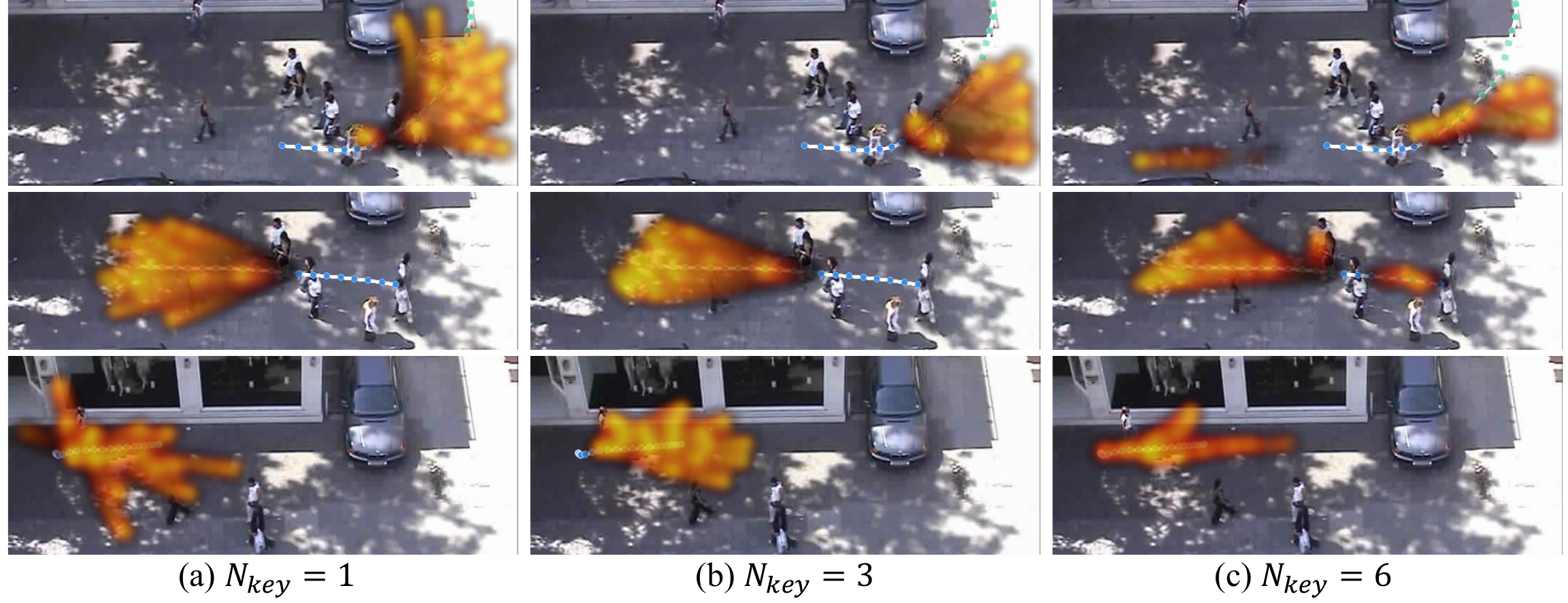}
    \caption{
        Number of Spectrum Keypoints and Prediction Styles.
        We show the visualized predictions with different $N_{key}$ configurations.
        Each sample has 20 random predictions.
    }
    \label{fig_number_keypoints}
\end{figure}

%% file: contents/_fig_vis.tex
\begin{figure*}[tb]
    \centering
    \includegraphics[width=1.0\linewidth]{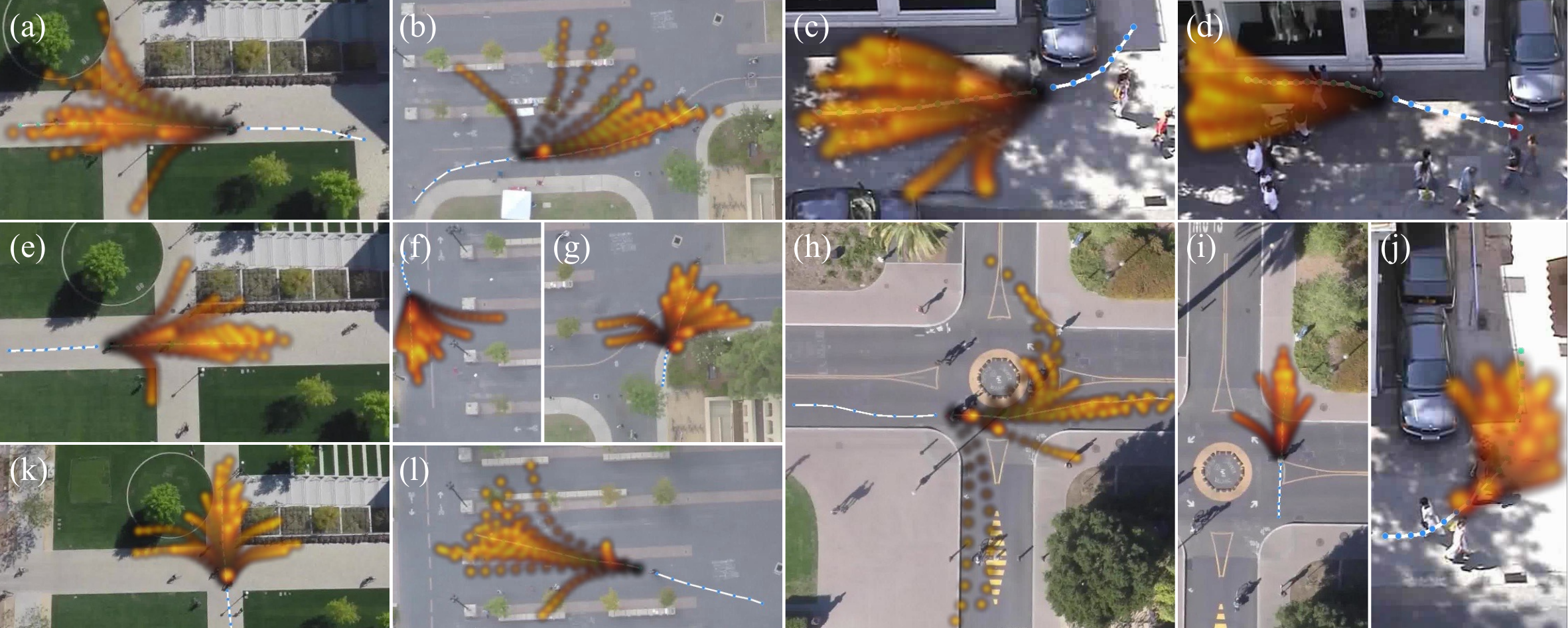}
    \caption{
        Visualized Predictions.
        Each sample illustrates 20 random predictions.
    }
    \label{fig_vis}
\end{figure*}

%% file: contents/_fig_failure.tex
\begin{figure}[tb]
    \centering
    \includegraphics[width=0.9\linewidth]{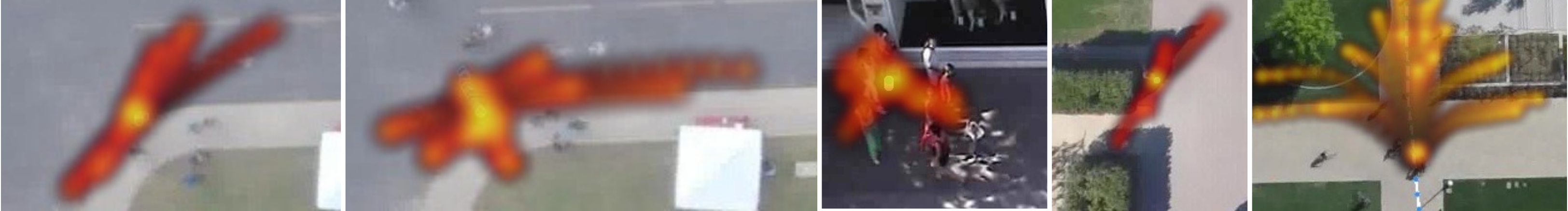}
    \caption{
        Failure Prediction Cases.
    }
    \label{fig_failure}
\end{figure}

%% file: contents/appendix.tex
\newpage
\appendix
\section{Detailed Transformer Structures}

We employ the Transformer \cite{attentionIsAllYouNeed} as the backbone to encode trajectory spectrums and the scene context in the two sub-networks.
The Transformer used in the \MODEL~has two main parts, the Transformer Encoder and the Transformer Decoder.
Both these two components are made up of several attention layers.

\SUBTITLE{Attention Layers}

Multi-Head Attention operations are applied in each of the attention layers.
Following definitions in \cite{attentionIsAllYouNeed}, each layer's multi-head dot product attention with $H$ heads is calculated as:
\begin{equation}
    \begin{aligned}
        \mbox{Attention}(q, k, v) & = \mbox{softmax}\left(\frac{qk^T}{\sqrt{d}}\right)v,                                         \\
        \mbox{MultiHead}(q, k, v) & = \mbox{fc}\left(\mbox{concat}(\left\{ \mbox{Attention}_i(q, k, v) \right\}_{i=1}^H)\right).
    \end{aligned}
\end{equation}
Here, $\mbox{fc}()$ denotes one fully connected layer that concatenates all heads' outputs.
Query matrix $q$, key matrix $k$, and value matrix $v$, are the three layer inputs.
Each attention layer also contains an MLP (denoted as MLP$_a$) to extract the attention features further.
It contains two fully connected layers.
ReLU activations are applied in the first layer.
Formally, we have the layer output $f_o$:
\begin{equation}
    f_{o} = \mbox{ATT}(q, k, v) = \mbox{MLP}_a(\mbox{MultiHead}(q, k, v)).
\end{equation}

\SUBTITLE{Transformer Encoder}

The transformer encoder comprises several encoder layers, and each encoder layer contains an attention layer and an encoder MLP (MLP$_e$).
Residual connections and normalization layers are applied to prevent the network from overfitting.
Let $h^{(l+1)}$ denote the output of $l$-th encoder layer, and $h^{(0)}$ denote the encoder's initial input.
For $l$-th encoder layer, we have
\begin{equation}
    \label{eq_alpha_encoder}
    \begin{aligned}
        a^{(l)}   & = \mbox{ATT}(h^{(l)}, h^{(l)}, h^{(l)}) + h^{(l)}, \\
        a^{(l)}_n & = \mbox{Normalization}(a^{(l)}),                   \\
        c^{(l)}   & = \mbox{MLP}_e(a_n^{(l)}) + a_n^{(l)},             \\
        h^{(l+1)} & = \mbox{Normalization}(c^{(l)}).
    \end{aligned}
\end{equation}

\SUBTITLE{Transformer Decoder}

Like the Transformer encoder, the Transformer decoder comprises several decoder layers, and each is stacked with two different attention layers.
The first attention layer in the Transformer decoder focuses on the essential parts in the Transformer encoder's outputs $h_e$ queried by the decoder's input $X$.
The second layer is the same self-attention layer as in the encoder.
Similar to \EQUA{eq_alpha_encoder}, we have:
\begin{equation}
    \label{eq_alpha_decoder}
    \begin{aligned}
        a^{(l)}      & = \mbox{ATT}(h^{(l)}, h^{(l)}, h^{(l)}) + h^{(l)}, \\
        a^{(l)}_n    & = \mbox{Normalization}(a^{(l)}),                   \\
        a_2^{(l)}    & = \mbox{ATT}(h_e, h^{(l)}, h^{(l)}) + h^{(l)},     \\
        a_{2n}^{(l)} & = \mbox{Normalization}(a_2^{(l)})                  \\
        c^{(l)}      & = \mbox{MLP}_d(a_{2n}^{(l)}) + a_{2n}^{(l)},       \\
        h^{(l+1)}    & = \mbox{Normalization}(c^{(l)}).
    \end{aligned}
\end{equation}

\SUBTITLE{Positional Encoding}

Before inputting agents' representations or trajectory spectrums into the Transformer, we add the positional coding to inform the relative position of each timestep or frequency portion in the sequential inputs.
The position coding $f_e^t$ at step $t~(1 \leq t \leq t_h)$ is obtained by:
\begin{equation}
    \begin{aligned}
        f_e^t                  & = \left({f_e^t}_0, ..., {f_e^t}_i, ..., {f_e^t}_{d-1}\right) \in \mathbb{R}^{d}, \\
        \mbox{where}~{f_e^t}_i & = \left\{\begin{aligned}
             & \sin \left(t / 10000^{d/i}\right),     & i \mbox{ is even}; \\
             & \cos \left(t / 10000^{d/(i-1)}\right), & i \mbox{ is odd}.
        \end{aligned}\right.
    \end{aligned}
\end{equation}
Then, we have the positional coding matrix $f_e$ that describes $t_h$ steps of sequences:
\begin{equation}
    f_e = (f_e^1, f_e^2, ..., f_e^{t_h})^T \in \mathbb{R}^{{t_h}\times d}.
\end{equation}
The final Transformer input $X_T$ is the addition of the original sequential input $X$ and the positional coding matrix $f_e$.
Formally,
\begin{equation}
    X_T = X + f_e \in \mathbb{R}^{t_h \times d}.
\end{equation}

\SUBTITLE{Layer Configurations}

We employ $L = 4$ layers of encoder-decoder structure with $H = 8$ attention heads in each Transformer-based sub-networks.
The MLP$_e$ and the MLP$_d$ have the same shape.
Both of them consist of two fully connected layers.
The first layer has 512 output units with the ReLU activation, and the second layer has 128 but does not use any activations.
The output dimensions of fully connected layers used in multi-head attention layers are set to $d$ = 128.

\section{Additional Analyses}

\subfile{./_tab_k.tex}

\SUBTITLE{Spectrums and Stochastic Performance}

Similar to previous multiple outputs approaches, the proposed \MODEL~could give multiple stochastic predictions to the same target agent.
More and more researchers \cite{mangalam2020not,mangalam2020s} have focused on their models' stochastic performance by adding the number of random generations.
Following these methods, we design ablation studies to verify \MODEL's stochastic performance as well as the performance improvement by using spectrums to model and forecast trajectories when generating a large number of stochastic predictions.
Detailed comparisons (\emph{best-of-k}) are shown in \TABLE{tab_k}.

Results show that describing and predicting trajectories with trajectory spectrums could bring even more significant performance gains than the original time series when generating more stochastic predictions.
The \MODEL~outperforms the other \MODEL~variation, which implements trajectories as time-series rather than spectrums, for about 12.3\% ADE and 24.8\% FDE when generating $k=100$ predictions on ETH-UCY.
It is worth noting that this performance gain amounts to the astonishing 50\% on \FDE~and 30\% \ADE~when outputting 1000 trajectories.
Results in SDD also present the same trends.
The performance gain has been improved from 0.5\% to 17.5\% on ADE and from 12.9\% to 51.2\% on FDE with the number of predictions raised to 1000.
Quantitative metrics illustrate the effectiveness of the \MODEL~in generating a large number of predictions.
On the other hand, it also demonstrates the impressive performance gains achieved by hierarchically predicting agents' trajectory spectrums.
Furthermore, it also shows the potential performance advantage of the proposed model in specific scenarios requiring a large number of multimodal predictions, such as trajectory prediction for traffic or autonomous driving.

In addition, the FDE will drop faster than the ADE as the generation number $k$ increases.
Using spectrums to predict trajectories in a hierarchical manner means that agents' overall motion trends (\IE, the keypoint spectrums) may obtain more substantial constraints, resulting in a fast decreasing trend of FDE.
Interestingly, the \PECNET~\cite{mangalam2020not} that divides the prediction pipeline into the ``destination prediction'' and the ``destination-conditioned prediction'' has a similar experimental phenomenon.
It also demonstrates the effectiveness of using a hierarchical strategy to predict agents' trajectories, although the two approaches use completely different perspectives to describe and predict trajectories.

\SUBTITLE{Smoothness}

In earlier studies on trajectory prediction, researchers have widely used \emph{autoregressive models}, such as recurrent neural network (RNN)\cite{structuralRNN} and long-short term memory (LSTM)\cite{socialLSTM}, to forcast agents' future trajectories \emph{recurrently}.
They model agents' behaviors and interactions in the next time step based on their last observed or predicted states.
These models could better describe the state changes between the adjacent moments.
Therefore, their predicted trajectories could reflect better smoothness characteristics.
However, more and more \emph{non-autoregressive models} like \cite{mangalam2020not,giuliari2020transformer,wong2021msn,mangalam2020s} are proposed and show excellent quantitative results.
These models predict agents' positions at different future moments ``in parallel'', which means that the relationship between the adjacent dots might be less concerned.

\FIG{fig_smooth}(a) (b) and (c) are visualized trajectories from \SOCIALGAN\cite{socialGAN}, \PECNET\cite{mangalam2020not}, and MSN\cite{wong2021msn}.
Although methods (b) and (c) have achieved better qualitative performance, some of their predictions do not seem so ``continuous''.
Specifically, some of these predicted trajectories present large fluctuations on adjacent time steps, while some predictions also exhibit abrupt state changes that may not happen under real-world physical constraints.
The proposed \MODEL~predicts agents' spectrums directly, rather than the time-sequences.
With the help of the DFT, \MODEL~could give ``smoother'' predictions than these models.
It means that in the IDFT process, the correlation between adjacent time steps will be smoothly constrained via sinusoids.
\FIG{fig_smooth}(d) and (e) show the relatively smooth predicted trajectories given by \MODEL.
It better reflects agents' real motion laws due to physical movement limitations.

It is worth noting that the mentioned ``smoothness'' does not mean that the smoother the network output is always the better.
Smoothness reflects the preferences and styles of agents' movements.
In current short-time (\IE, recording 3.2s' trajectory to forecast 4.8s' trajectory) trajectory prediction tasks (like the trajectory prediction mentioned in this paper), since the target agents (including pedestrians and bikers, etc.) usually do not move extensively, their observed trajectories tend to exhibit smooth properties as well.
In longer time predictions, maintaining their movement style will be a further counterpart of the mentioned smoothness, and it might be made more accessible by reflecting agents' movement preferences through the trajectory spectrums.

\subfile{./_fig_smooth.tex}

%% file: contents/_tab_k.tex
\begin{table}[tbph]
    \centering
    \caption{
        Verifying of Stochastic Performance.
        $k$ indicates the number of stochastic generations.
        Metrics are ``\ADE/\FDE'' under the \emph{best-of-k} validation.
        Lower is better.
    }
    \label{tab_k}

    \begin{tabular}{c|c||c|c||c|c}
        \hline
        Datasets & $k$ & \MODEL 
        & \makecell[c]{\MODEL\\(without DFT)} 
        & \makecell[c]{Absolute\\Improvements} 
        & \makecell[c]{Percentage\\Improvements} \\
        
        \hline
        \multirow{5}{*}{\rotatebox{90}{\makecell[c]{ETH-UCY\\(Average)}}}
        & 100 & 0.14/0.17 & 0.16/0.23 & 0.02/0.06 & 12.3\% / 24.8\% \\
        & 200 & 0.12/0.12 & 0.15/0.20 & 0.03/0.08 & 20.1\% / 34.6\% \\
        & 400 & 0.10/0.09 & 0.14/0.17 & 0.04/0.08 & 24.5\% / 44.1\% \\
        & 600 & 0.10/0.07 & 0.13/0.15 & 0.03/0.08 & 26.8\% / 46.7\% \\
        & 1000 & 0.09/0.06 & 0.13/0.13 & 0.04/0.07 & 31.4\% / 51.2\% \\

        \hline
        \multirow{5}{*}{\rotatebox{90}{SDD}}
        & 100 & 5.80/6.90 & 5.83/7.93 & 0.03/1.03 & 0.5\% / 12.9\% \\
        & 200 & 4.95/4.95 & 5.27/6.49 & 0.32/1.54 & 6.0\% / 23.6\% \\
        & 400 & 4.30/3.52 & 4.86/5.30 & 0.56/1.78 & 11.5\% / 33.6\% \\
        & 600 & 3.98/2.87 & 4.68/4.71 & 0.70/1.84 & 14.8\% / 38.9\% \\
        & 1000 & 3.67/2.28 & 4.46/4.03 & 0.79/1.75 & 17.5\% / 43.4\% \\

        \hline
    \end{tabular}
\end{table}

%% file: contents/_fig_smooth.tex
\begin{figure}[tbp]
    \centering
    \includegraphics[width=1.0\linewidth]{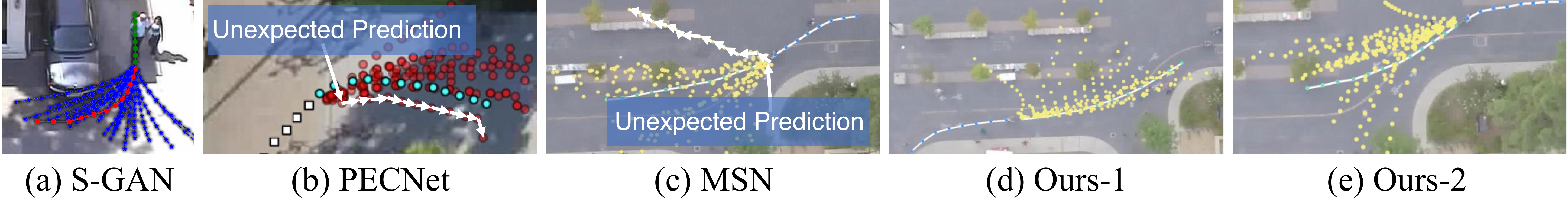}
    \caption{
        Smoothness Comparisons.
        (a), (b) and (c) are visualized predictions from \SOCIALGAN\cite{socialGAN}, \PECNET\cite{mangalam2020not}, and MSN\cite{wong2021msn}.
        (d) and (e) are predictions given by our \MODEL.
        With the help of spectrums, \MODEL's predictions show better continuous and smoothness.
    }
    \label{fig_smooth}
\end{figure}